\documentclass{article}
\usepackage[final,nonatbib]{nips_2016}

\usepackage[utf8]{inputenc} 
\usepackage[T1]{fontenc}    
\usepackage{hyperref}       
\usepackage{url}            
\usepackage{booktabs}       
\usepackage{amsfonts}       
\usepackage{nicefrac}       
\usepackage{microtype}      

\usepackage{algorithm}
\usepackage{algorithmic}
\usepackage{xspace,wrapfig}
\usepackage{color}
\usepackage{subfigure}
\usepackage{bbm}
\usepackage{makeidx}  
\usepackage{times,amsmath,amssymb,amsthm,graphicx,url,xspace,wrapfig}

\usepackage{url}
\usepackage{xcolor,graphicx,xkeyval,calc,tikz,ifthen}
\usepackage{comment}
\usepackage{enumitem}
\usepackage{caption}
\usepackage{enumitem} 
\usepackage{wrapfig}
\usepackage{hyperref}

\newcommand{\POIM}{\textrm{POIM}\xspace}
\newcommand{\SATRIX}{\textrm{MFI}\xspace}
\newcommand{\kernelSATRIX}{\textrm{kernel MFI}\xspace}
\newtheorem{mylemma}{Lemma}
\newtheorem{mydef}{Definition}

\newcommand{\Ycal}{{\mathcal{Y}}}

\title{Feature Importance Measure for Non-linear Learning Algorithms}

\author{
	Marina M.-C. Vidovic\\
	Machine Learning Group\\
	Technical University of Berlin\\
	Berlin, Germany \\
	\texttt{marina.vidovic@tu-berlin.de}\And
	Nico G\"ornitz\\
	Machine Learning Group\\
	Technical University of Berlin\\
	Berlin, Germany \\
	\texttt{nico.goernitz@tu-berlin.de} \And
	Klaus-Robert M\"uller\\
	Machine Learning Group\\
	Technical University of Berlin\\
	Berlin, Germany \\
	\texttt{klaus-robert.mueller@tu-berlin.de} \And
	Marius Kloft\\
	Department of Computer Science\\
	Humbold University of Berlin\\
	Berlin, Germany \\
	\texttt{kloft@hu-berlin.de} }
\begin{document}
	\maketitle
	\section{Introduction}
    \vspace{-0.1cm}
	\noindent Complex problems may require sophisticated, non-linear learning methods
	such as kernel machines or deep neural networks to achieve state of the art 
	prediction accuracies. However, high prediction accuracies are not the only objective to consider 
	when solving problems using machine learning. Instead, particular scientific applications require some explanation of the learned prediction function. 
     Unfortunately, most methods 
    do not come with out of the box straight forward interpretation.
	Even linear prediction functions $s(x)=\sum_j \beta_j x_j$ are \emph{not} straight forward to explain if
    features $\beta$ exhibit complex correlation structure.
	
	
	In computational biology, positional oligomer importance matrices (POIMs) \cite{SonZiePhiRae08} address the need for interpretation of sophisticated learning machines.
	POIMS specifically explain the output of kernel-based learning methods acting on DNA sequences using a 
	weighted degree string kernel \cite{BenOngSonSchRae08,SonSchPhiBehRae07,SchSmo2002,RaeSonSriWitMueSomSch07}.
	A WD kernel breaks two discrete DNA sequences $x$ and $x'$ of length $L$ apart into all subsequences 
	up to some length and then counts the number of matching subsequences---the so-called \emph{positional oligomers} (POs). 
    For the following considerations, let $\Sigma=\{A,C,G,T\}$ be the DNA alphabet and $X\in \Sigma^L$ a random variable over the DNA alphabet of length $L$. 
	POIMs assign each PO $y\in \Sigma^k$, of length $k$ starting at position $j$ in $X$ with an importance score $\POIM_{y,j} \propto \mathbb{E}[s(X)|X_{j:j+k}=y]$. 
	POIMs allow visualization of each PO's significance to the prediction function $s$. 
	A seminal property of POIMs is that they take the overlaps of the POs at different positions and lengths into account.
		As visual inspecting POIMs can be tedious, \cite{VidGoeMueRaeKlo2015b,VidGoeMueRaeKlo2015} proposed motifPOIMs, 
	a probabilistic non-convex method to automatically extract the biological factors underlying the SVM's prediction such as promoter elements or transcription factor binding sites --often called \textrm{motifs}.
    Unfortunately, POIMs are restricted to specific DNA applications.
    
	As a generalization of POIMs,  the feature importance ranking measure (FIRM) \cite{Zien2009} assigns each feature f with an importance score 
	$ Q_f := \sqrt{{\rm Var}_Y[\mathbb{E}_{X}[s(X)| f(X)=Y]}\thinspace.$
	FIRM measures the variation of the prediction function when varying a feature.
	If the expected value of the prediction function is not changed when varying a feature f,
	the feature is considered as unimportant.
	Unfortunately, FIRM is in general intractable~\cite{Zien2009}.
	
	In this paper, we propose the \emph{Measure of Feature Importance} (\SATRIX).
	\SATRIX is general and can be applied to any arbitrary learning machine (including kernel machines and deep learning).
	\SATRIX is intrinsically non-linear and can detect features that by itself are inconspicuous and only impact the prediction function through their interaction with other features.
	Lastly, \SATRIX can be used for both --- model-based feature importance (as POIMs and FIRM) and instance-based 
	feature importance (i.e, measuring the importance of a feature for a particular data point).
	
	\section{Methodology}
	In this section, we describe our proposed method --- Measure of Feature Importance (\SATRIX).
	\SATRIX extends the concepts of POIM and FIRM (which are contained as special cases) to non-linear feature 
	interactions and instance-based feature importance attribution, and it is particularly simple to apply. To distinguish between model-based and instance-based \SATRIX, we introduce a function called ``explanation mode'', which maps the sample in their respective feature space. Exemplary, for instance-based explanation, a DNA sequence would be mapped to itself, whereas the same sequence would be mapped to a POIM in case of model-based explanation.	
	\begin{mydef}[\SATRIX and \kernelSATRIX]\label{def:satrix}
		Let $X$ be a random variable on a space $\mathcal X$. Furthermore, let $s: \mathcal{X} \rightarrow \mathcal{Y}$ be a prediction function (output by an arbitrary learning machine),
		and let $f: \mathcal{X} \rightarrow \mathbb{R}$ be a real-valued feature.
		Let $\phi: \mathcal{X} \rightarrow F$ be a function (``explanation mode''), where $F$ is an arbitrary space.
		Lastly, let $k: \Ycal \times \Ycal \rightarrow \mathbb{R}$ and $l: \mathbb{R} \times \mathbb{R} \rightarrow \mathbb{R}$ be kernel functions. 
		Then we define:
		\begin{eqnarray}
			\SATRIX: & S_{\phi,f}(t) := \mathbb{E}[s(X)\phi(X)|f(X)=t] \label{eq:mfi}\\
			\text{kernel}  \ \SATRIX: & S^+_{\phi,f}(t) = Cov[k(s(X), s(\cdot)), l(\phi(X), \phi(\cdot))|f(X)=t]. \label{eq:kernel mfi}
		\end{eqnarray}
	\end{mydef}
	\begin{table}[!ht]
		\centering
		\caption{\textbf{Specific instantiation for \SATRIX} in terms of instance-based (ib) and model-based (mb) application, with $B\in\mathcal{A}^k \times \{1,\ldots,L-k+1\}$\thinspace . Illustrations are given in Figure \ref{fig:gpoim_examples}.}
        \begin{small}
		{\begin{tabular}{|c|l|c|c|ccc|}\hline
				& Objects &mode&method& $\phi$ & $f$ & t\\
				\hline
				\hline
				a & Image z &ib& \SATRIX & $\phi(X)=1$ & $f_{i,j}(X)=X_{i,j}$ & $t=z_{i,j}$\\
				b & Sequence z &ib& \SATRIX &$\phi(X)=1$ & $f_{i,k}(X)=X_{i:i+k}$ & $t=z_{i:i+k}$ \\
				\hline
				c & Images &mb& \kernelSATRIX & $\phi(X)=X$ & $f(X)=t$ & $const$ \\
				d & Sequences &mb& \kernelSATRIX &$\phi(X)=B$ & $f(X)=t$ & $const$\\
				\hline
			\end{tabular}}
            \end{small}
			\label{table:gpoim-settings}
		\end{table}
		\begin{figure*}[h!]
			\centering
			\includegraphics[width=0.9\textwidth]{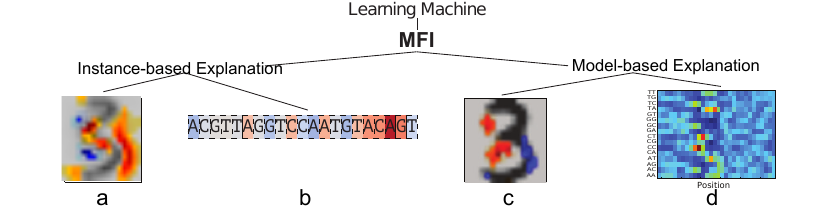}
			\vspace{-0.2cm}	\caption{\label{fig:gpoim_examples}\textbf{\SATRIX Examples} We consider two possible flavors of feature importance: (left) instance-based importance measures (e.g. \emph{Why is this specific example of '3' classified as '3' using my trained RBF-SVM classifier?}); (right) model-based importance measure (e.g. \emph{Which regions are generally important for the classifier decision?}). }
			\vspace{-0.2cm}
		\end{figure*}
		
		In the following, we will explain the ``explanation mode'' of above definition in terms of model-based and instance-based proceed exemplary for both, sequence and image data. 
		\subsection{Model-based \SATRIX:} 
        Here, the task is to globally assess what features a given (trained) learning machine regards as most significant --- independent of the examples given. In the case of sequence data, were we have sequences of length $L$ over the alphabet $\Sigma=\{A,C,G,T\}$, an importance map for all $k$-mers over all positions is gained by using the explanation mode $\phi: \Sigma^L \rightarrow \Sigma^{k\times L-k+1}$, where each sequence is mapped to a sparse PWM, in which entries only indicate presence or absence of positional $k$-mers. 
		In the case of two dimensional image data, $X \in \mathbb{R}^{d_1 \times d_2}$, where we already are in the decent visual explanation mode, $\phi(X)=X$ keeps the surroundings by mapping the data to itself. 
		In both cases, we set $f(X)=t$, where $t=const$, which is why we can neglected it. The various case studies are summarized in Table \ref{table:gpoim-settings} with corresponding examples shown in Figure \ref{fig:gpoim_examples} on the right.
		\subsection{Instance-based \SATRIX:} 
        Given a specific example, the task at hand is to assess why this example has been assigned this specific classifier score (or class) prediction. In the case of sequence data we compute the feature importance of any positional $k$-mer in a given sequence $g\in \Sigma^L$ by $f(X)=X_{i:i+k}$, with $t=g_{i:i+k}$.
		In the case of images, where $g\in \mathbb{R}^{d_1\times d_2}$ is the image of interest and $g_{i,j}$ expose one pixel, $f(X)=X_{i,j}$ maps the random samples $X\in \mathbb{R}^{d_1\times d_2}$ to one pixel $t=g_{i,j}$.
		In both cases, we set $\phi(X)=1$, which is why we can neglect it. For examples and specific instruction see Table \ref{table:gpoim-settings} and Figure \ref{fig:gpoim_examples} on the left.

		\subsection{Relation to Hilbert-Schmidt Independence Criterion}
		In 2005, the Hilbert-Schmidt independence criterion~\cite{GreBouSmoSch05} (HSIC) was 
		proposed as a kernel-based methodology to measure the independence of two distinct variables $X$ and $Y$:
		\begin{align*}
			&HSIC(X,Y) = \|C_{XY} \|^2 = \mathbb{E}[k(X,X')l(Y,Y')] \\
			& -2\mathbb{E}[\mathbb{E}_X[k(X,X')]\mathbb{E}_Y[l(Y,Y')]] + \mathbb{E}[k(X,X')]\mathbb{E}[l(Y,Y')]
		\end{align*}
		where $k$ and $l$ are reproducing kernels and $C_{XY}$ is the cross-covariance operator.
		We have the following interesting relation of \SATRIX to HSIC.

		\begin{mylemma}[Relation of Kernel \SATRIX to HSIC]\label{def:hsic-relation}
			Given the \text{kernel} \SATRIX of Definition \ref{def:satrix} $S^+_{\phi,f}$, then $S^+_{\phi,f} = Cov[k(s(X), s(\cdot)), l(\phi(X), \phi(\cdot))|f(X)=t]$
			and the corresponding Hilbert-Schmidt Independence Criterion becomes:  $HSIC(S_{\phi}, \Ycal, \mathbb{R}) = \| S^+_{\phi} \|^2 = tr(KL) \thinspace .$
		\end{mylemma}
		The relation to HSIC provides us with a practical tool to assess non-linear feature importances as defined in kernel MFI in Definition \ref{def:satrix}.
		%
		%
		In order to make this approach practically suitable, we resort to sampling as an inference method. 
		To this end, let $Z\subset \mathcal{X}$ be a subset of $\mathcal{X}$ containing $n=|Z|$ samples.
		Then Eq.~\eqref{eq:mfi} can be approximated by
		$\hat{S}_{\phi,f}(t):=\frac{1}{|Z_{\{f(z)=t\}}|}\sum_{z\in Z}s(z)\phi(z)\mathbf{1}_{\{f(z)=t\}} - \mu_s \mu_{\phi}$
		with $\mu_{\phi} = \frac{1}{ |Z_{\{f(z)=t\}}| } \sum_{z \in Z_{\{f(z)=t\}}} \phi(z)$
		and $\mu_s = \frac{1}{ |Z_{\{f(z)=t\}}| } \sum_{z \in Z_{\{f(z)=t\}}} s(z)$.
		Hence, when number of samples $|Z| \rightarrow \infty$, then $\hat{S}_{\phi,f} \rightarrow S_{\phi,f}$. A corresponding sampling scheme is also available for kernel \SATRIX. 
		\section{Empirical Evaluation}
		\begin{wrapfigure}{r}{0.45\textwidth} 
			\vspace{-0.5cm}
			\includegraphics[width=6.0cm]{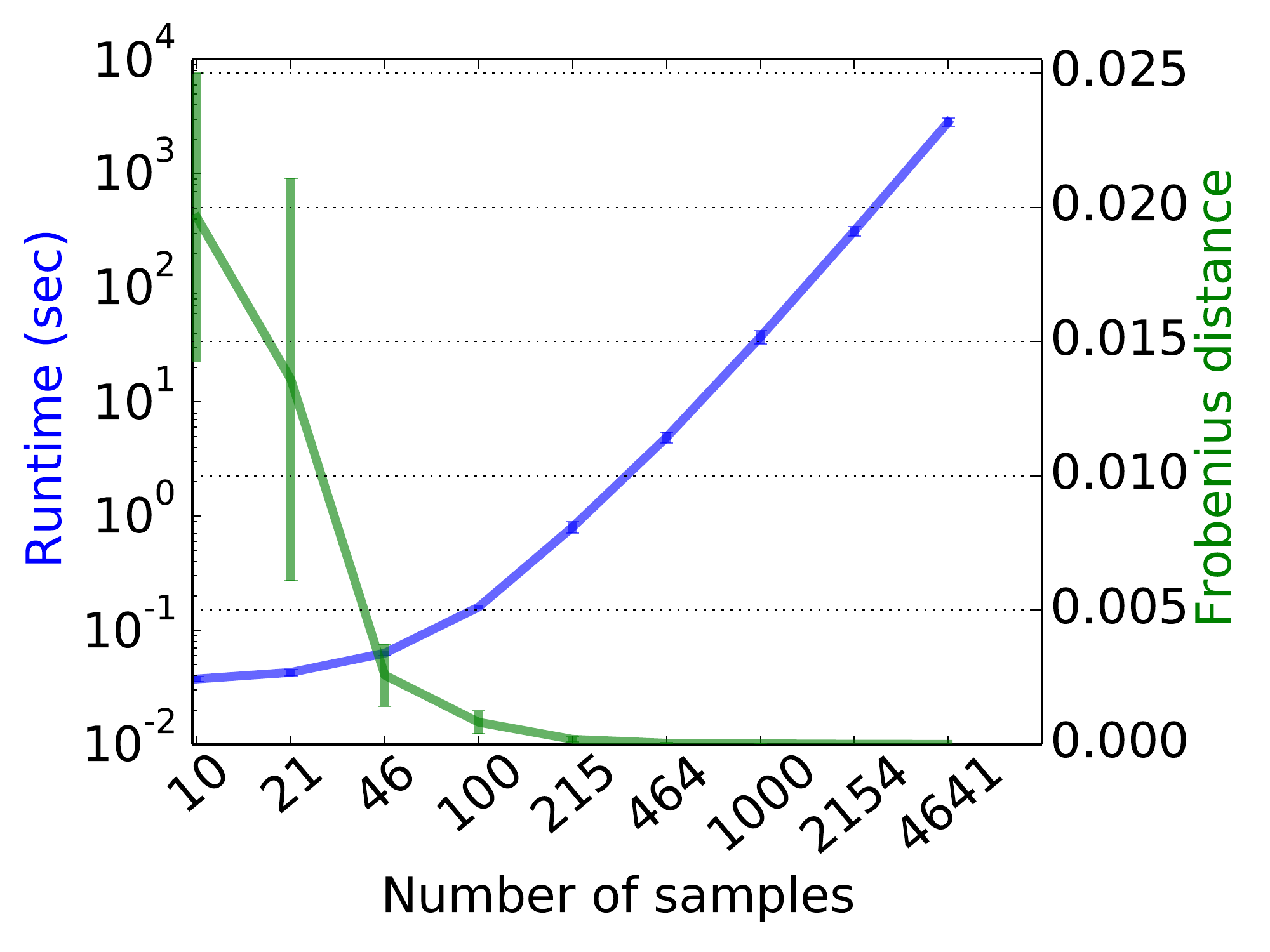}
			\vspace{-0.3cm}	\caption{\label{fig:runtime_motifdiff}{\bf Illustration of the runtime} measured in seconds for various sample sizes (plotted in blue) and of the Frobenius distance between two consecutive results (green curve).}
			\vspace{-1.0cm}
		\end{wrapfigure}
        In this section, we evaluate the proposed method empirically regarding its ability to explain the relevance of features for model- and instance-based explanation models. Although our method can be applied to any learning machines,
		we focus in the experiments on support vector machines (SVMs) using a Gaussian kernel function and convolutional neural networks (CNNs).
		\subsection{Experimental setup}
        \vspace{-0.1cm}
		For validation, we follow the Most Relevant First (MoRF) strategy \cite{samek2015evaluating} and successively calculate the classifier performance 
		while blurring pixels a) with descending relevance (i.e., computed by our proposed method) and b) randomly. The idea is that blurring pixels with high relevance will influence the classifier decision and thus drop its performance faster 
		than blurring randomly chosen pixels would do. 
		In the following, we evaluate our proposed method on the USPS data set, using a SVM with an
		RBF kernel and a CNN with following architecture: a 2D convolution layer with 10 tanh-filters of size 8x4, a max-pool layer of size=2, a dense-layer with 100 ReLUs, a dense layer with 2 softmax units. For all experiments, we used a sample size of 1000 samples, which was considered as suitable trade-off between runtime and accuracy.
		\begin{figure}[h!]
			\vspace{-0.4cm}
			\centering
			\subfigure[SVM]{\includegraphics[width=2.7cm]
				{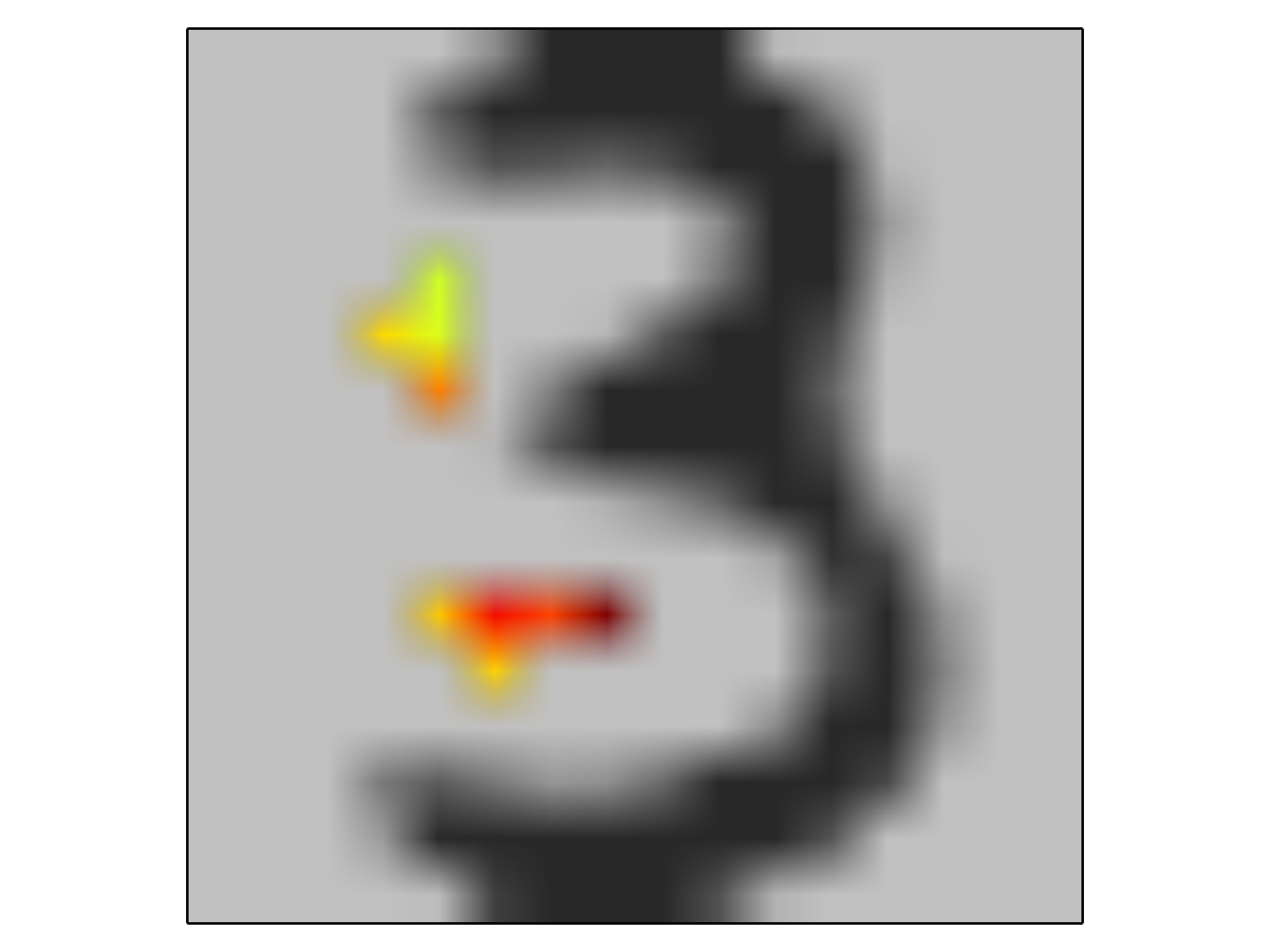}}
			\subfigure[CNN]{\includegraphics[width=2.7cm]{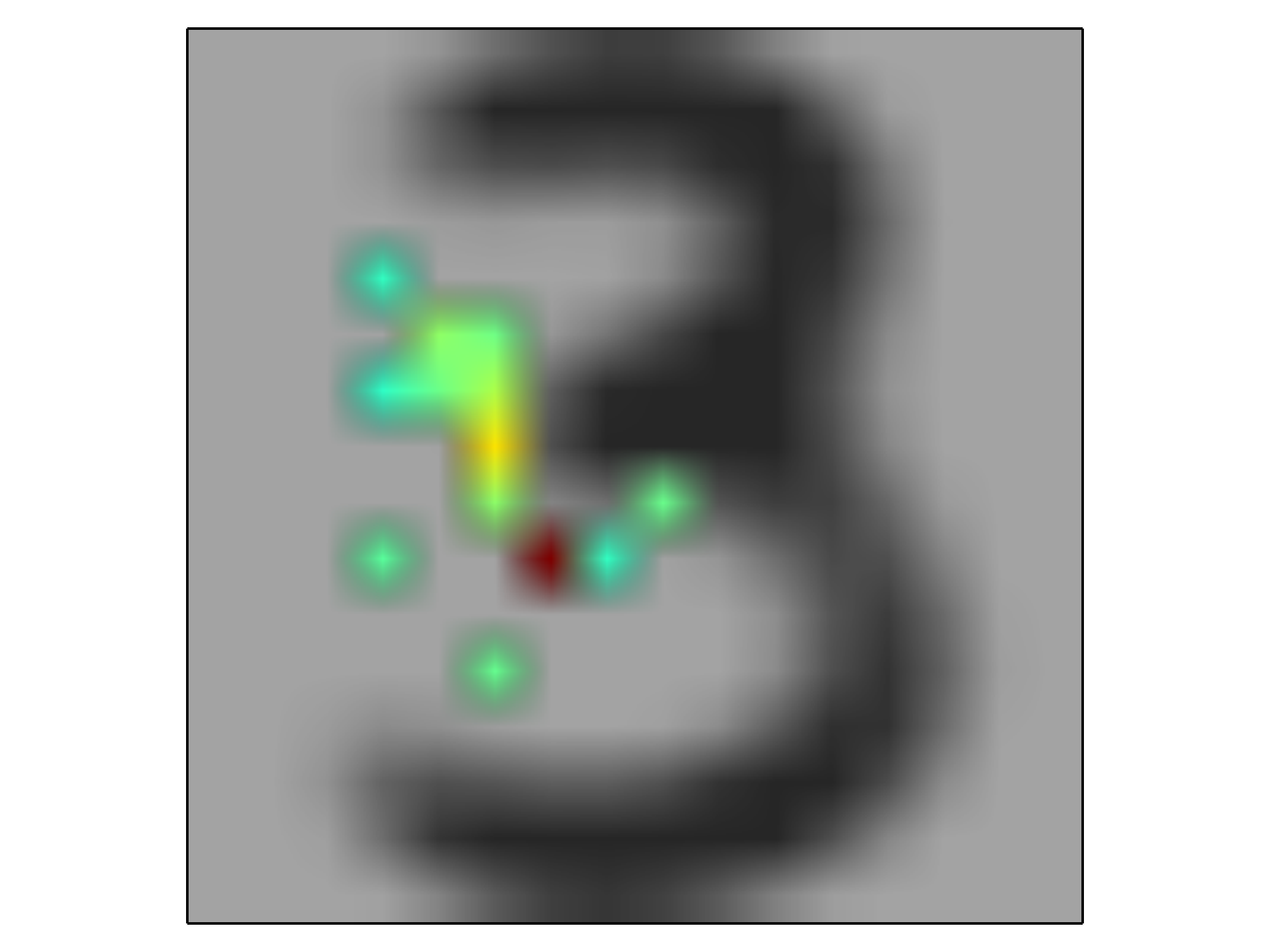}}
			\subfigure[SVM]{\includegraphics[width=2.9cm]{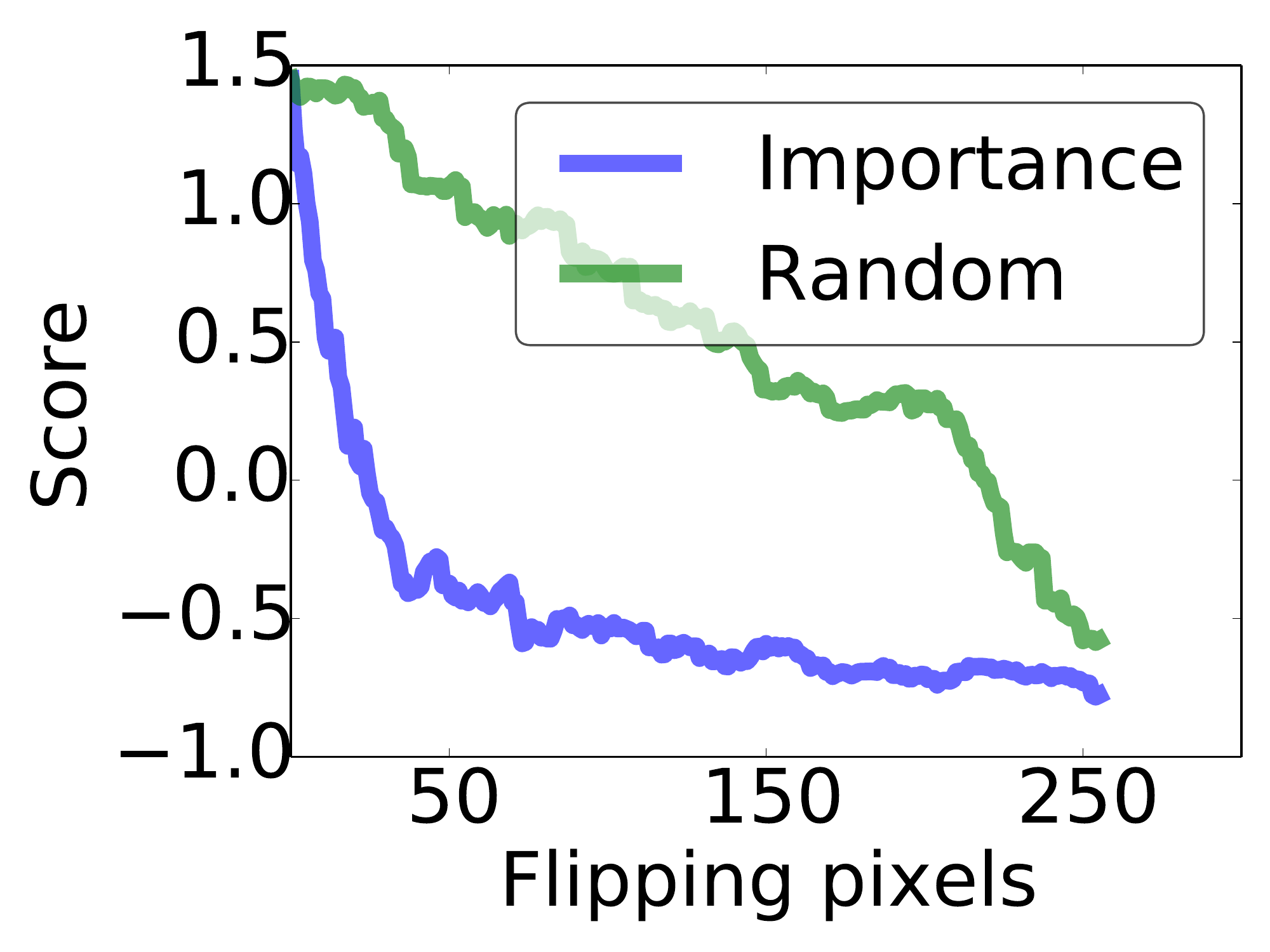}}
			\subfigure[CNN]{\includegraphics[width=2.9cm]{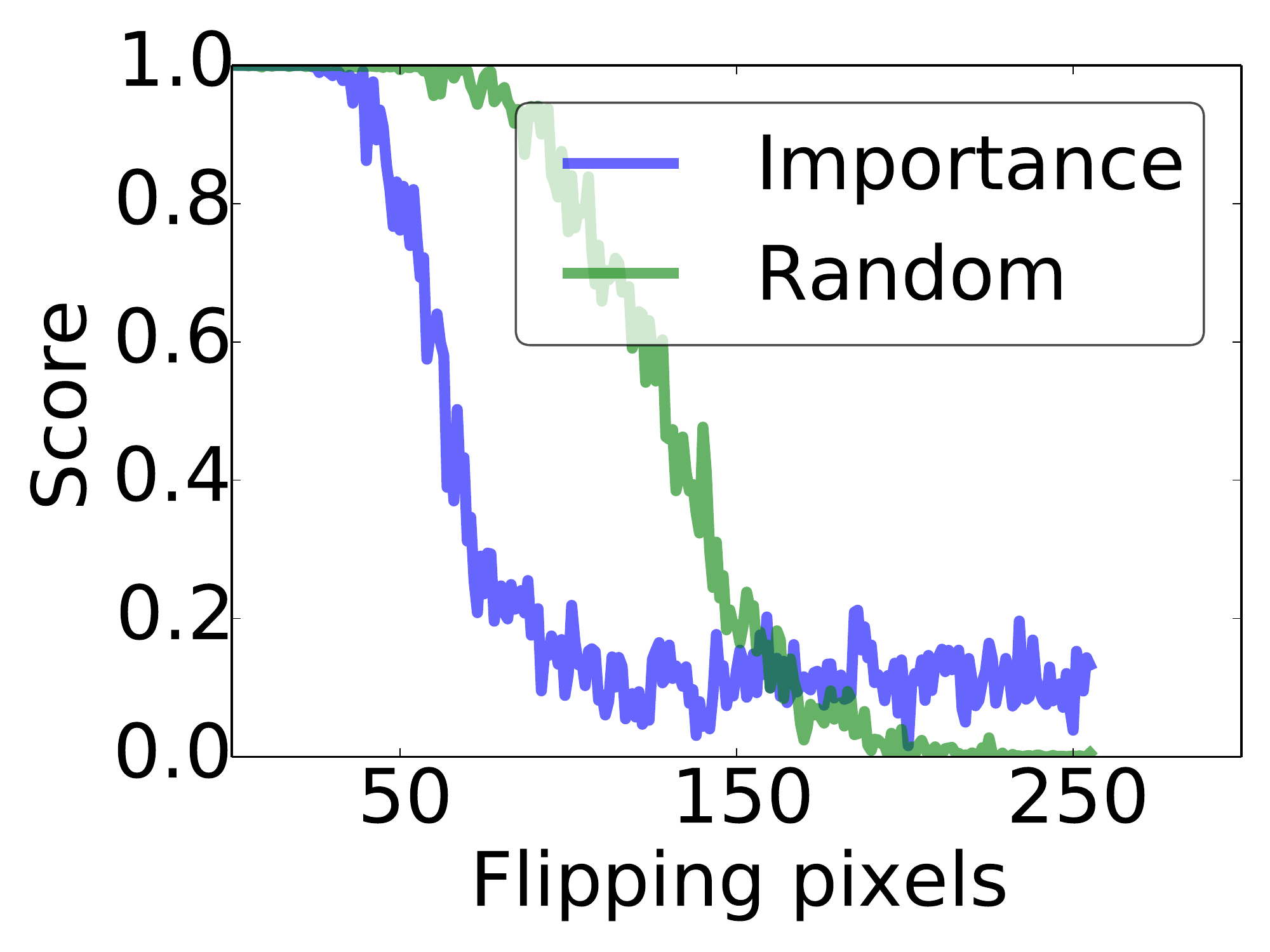}}
			\caption{\label{fig:USPSexperimentModelWise} \textbf{Results are shown for the USPS data set using \kernelSATRIX for SVM (a) and CNN (b)}, where the most important pixels found by \kernelSATRIX are embedded in the mean picture of digit \texttt{three}. 				Figure (c) and (d) show the classifier performance loss when successively blurring the pixel regarding their relevance found by \kernelSATRIX compared to a random pixel blurring.}
		\end{figure}
		\subsection{Results}
        To find a suitable trade-off between runtime and accuracy, we evaluate runtime and convergence behavior (in terms of the Frobenius distance of two consecutive results) for increasing numbers of samples. From the results, shown in Figure \ref{fig:runtime_motifdiff}, we observe that the Frobenius distance (green curve) converges to zero already for small sample sizes (215 samples). Unfortunately, runtime grows very fast (almost exponentially) showing the boundaries of our method. Hence, a good trade-off between runtime and accuracy would be any sample size between 500 and 2000 in this experiment. For the following experiments we used a sample size of 1000.

		\paragraph{Model-Based Feature Importance}
                \vspace{-0.1cm}
		
		The results are shown in Figure \ref{fig:USPSexperimentModelWise}.
		We observe that for both, SVM and CNN, the pixel bridge that changes the digit three to the digit eight is of high importance. 
		In Figure \ref{fig:USPSexperimentModelWise} (c) and (d) the classifier performance for increasing amount of blurring pixels in terms of MoRF 
		as explained above is shown. Compared to a random pixel blurring, we can clearly observe that the performance drops significantly faster 
		when blurring the most important pixels (as found by our proposed kernel \SATRIX method).
		\begin{wrapfigure}{r}{0.5\textwidth} 
			{\includegraphics[width=0.50\textwidth]{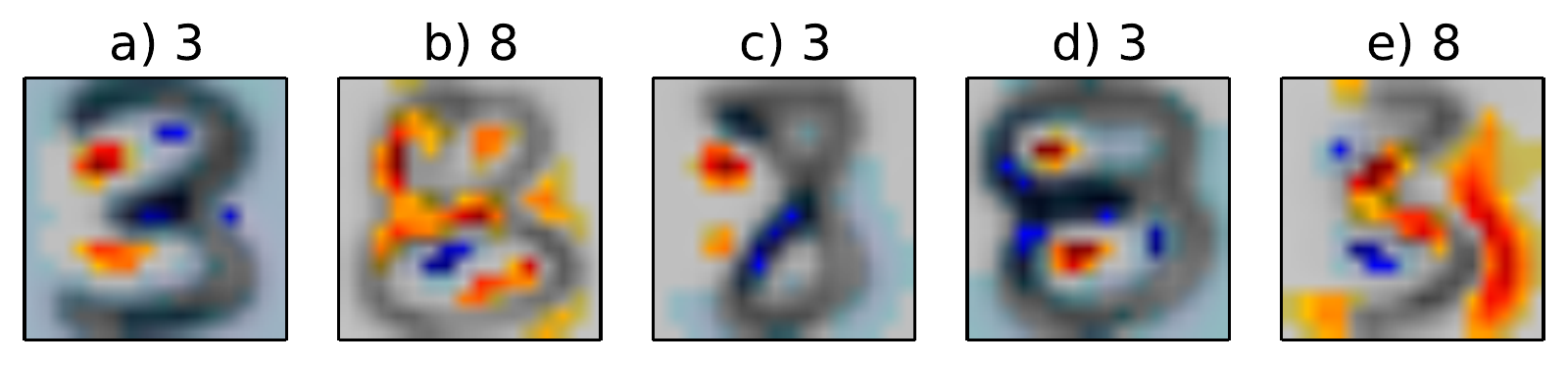}}
			\vspace{-0.5cm}
			\caption{\label{fig:test_grayscale} \textbf{Instance-based explanation of the SVM decision for five USPS test data images}. 
The highlighted pixels are informative for the individual SVM decisions (plotted at the image top) -- only the first two images were correctly classified.}
			\vspace{-0.8cm}
		\end{wrapfigure}
		\paragraph{Instance-Based Feature Importances}
		For the pixel-wise explanation experiment, an SVM with an RBF kernel was trained on the USPS training data set.
		From Figure \ref{fig:test_grayscale} we observe that the pixels building the vertical connection from a \texttt{three}  to an \texttt{eight} have a strong discriminative evidence. 
		If these positions are left blank, the image is classified as \texttt{three}, which, in case of the last three images leads to mis-classifications.   
		
\begin{figure*}[h!]	
        \centering
			{\includegraphics[width=0.77\textwidth]{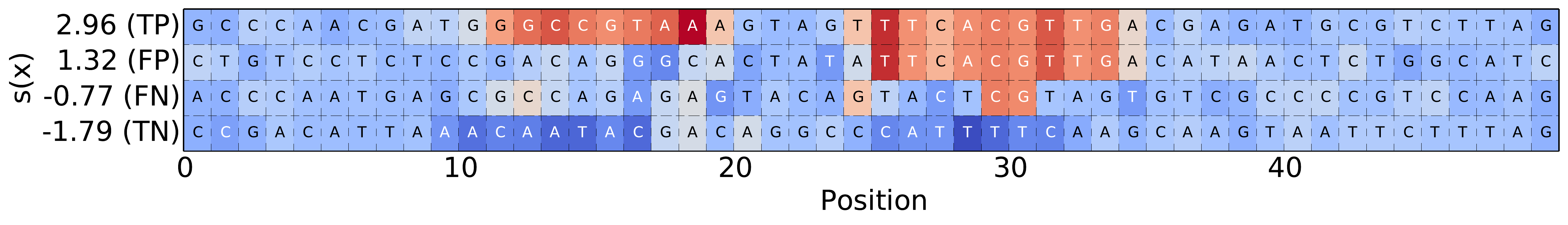}}
			\vspace*{-0.4cm}
			\caption{\label{fig:nucleotidwiseEplanation}{\bf Instance-based feature importances experiment.} 
				The highlighted nucleotids are informative for the SVM decision for four test sequences that have been correctly (TN and TP) and incorrectly (FP and FN) classified. 
}
			\vspace*{-0.6cm}
		\end{figure*}

		For the nucleotide-wise explanation experiment, an SVM with an WD kernel was trained on a synthetic training data set. 
		We inserted two motifs in the positive class (GGCCGTAAA at position 11 and TTTCACGTTGA at position 24).
		From Figure \ref{fig:nucleotidwiseEplanation} we observe that the nucleotides building the two patterns, 
		which we inserted in the positive sequences have strong discriminative evidence. 
		If the discriminative patterns are too noisy, the sequences are assumed to stem from the negative class, which, 
		in case of the false negative (FN) example leads to mis-classifications. 
		If only one of the two patterns was inserted, the classifier gives high evidence to the single pattern and assigns the wrong label. 
		\section{Conclusion \& Outlook}
		\vspace{-0.1cm}
		By this work, we contributed to opening the black box of learning machines. 
		Building up on POIMs and FIRM, we  proposed \SATRIX, which is a general measure of feature importance that is applicable to arbitrary learning machines.
		\SATRIX can be used for both for a general explanation of the prediction model and for a data instance specific explanation.
		As a nonlinear measure, \SATRIX can detect features that exhibit their importance only through interactions with other features.
		Experiments on artificially generated splice-site sequence data as well as real-world image data demonstrate the properties and benefits of our approach.\\
		While in the present work  we have focused images and sequences, the framework allows us to explain arbitrary data sources.
		In future research, we would like to study further applications (e.g., involving trees, graphs, etc),
		including  wind turbine anomaly detection, as well as we want to investigate advanced  sampling techniques 
		from probabilistic machine learning that may lead to faster convergence.
        
			\section*{Acknowledgments}
			
			MMCV and NG were supported by BMBF ALICE II grant 01IB15001B. 
			We also acknowledge the support by the German Research Foundation through the grant DFG KL2698/2-1, MU 987/6-1, and RA 1894/1-1. KRM thanks for partial funding by the National Research Foundation of Korea funded by the Ministry of Education, Science, and Technology in the BK21 program. MK and KRM were supported by the German Ministry for Education and Research through the awards 031L0023A and 031B0187B and the Berlin Big Data Center BBDC (01IS14013A). 
			
		\bibliographystyle{abbrv}
		\bibliography{poim_bib,new,Mendeley}

\begin{thebibliography}{10}

\bibitem{BenOngSonSchRae08}
A.~Ben-Hur, C.~S. Ong, S.~Sonnenburg, B.~Schoelkopf, and G.~Raetsch.
\newblock {Support vector machines and kernels for computational biology}.
\newblock {\em PLoS Computational Biology}, 4(10), 2008.

\bibitem{GreBouSmoSch05}
A.~Gretton, O.~Bousquet, A.~Smola, and B.~Sch{\"{o}}lkopf.
\newblock {Measuring Statistical Dependence with Hilbert-Schmidt Norms}.
\newblock In {\em International conference on algorithmic learning theory},
  2005.

\bibitem{RaeSonSriWitMueSomSch07}
G.~R\"atsch, S.~Sonnenburg, J.~Srinivasan, H.~Witte, K.~R. M\"uller, R.~J.
  Sommer, and B.~Schoelkopf.
\newblock {Improving the Caenorhabditis elegans genome annotation using machine
  learning}.
\newblock {\em PLoS Computational Biology}, 3(2):0313--0322, 2007.

\bibitem{samek2015evaluating}
W.~Samek, A.~Binder, G.~Montavon, S.~Bach, and K.-R. M{\"u}ller.
\newblock Evaluating the visualization of what a deep neural network has
  learned.
\newblock {\em arXiv preprint arXiv:1509.06321}, 2015.

\bibitem{SchSmo2002}
B.~Sch{\"{o}}lkopf and A.~J. Smola.
\newblock {\em {Learning with Kernels: Support Vector Machines, Regularization,
  Optimization, and Beyond}}.
\newblock MIT Press, 2002.

\bibitem{SonSchPhiBehRae07}
S.~Sonnenburg, G.~Schweikert, P.~Philips, J.~Behr, and G.~R{\"{a}}tsch.
\newblock {Accurate splice site prediction using support vector machines}.
\newblock {\em BMC Bioinformatics}, 8(Suppl 10):S7, 2007.

\bibitem{SonZiePhiRae08}
S.~Sonnenburg, A.~Zien, P.~Philips, and G.~R{\"{a}}tsch.
\newblock {POIMs: Positional oligomer importance matrices - Understanding
  support vector machine-based signal detectors}.
\newblock {\em Bioinformatics}, 24(13):6--14, 2008.

\bibitem{VidGoeMueRaeKlo2015}
M.~M.-C. Vidovic, N.~G{\"{o}}rnitz, K.-R. M{\"{u}}ller, G.~R{\"{a}}tsch, and
  M.~Kloft.
\newblock {Opening the Black Box: Revealing Interpretable Sequence Motifs in
  Kernel-Based Learning Algorithms}.
\newblock In {\em ECML PKDD}, volume 6913, pages 175--190, 2015.

\bibitem{VidGoeMueRaeKlo2015b}
M.~M.-C. Vidovic, N.~G{\"{o}}rnitz, K.-R. M{\"{u}}ller, G.~R{\"{a}}tsch, and
  M.~Kloft.
\newblock {SVM2Motif — Reconstructing Overlapping DNA Sequence Motifs by
  Mimicking an SVM Predictor}.
\newblock {\em PLoS ONE}, pages 1--23, 2015.

\bibitem{Zien2009}
A.~Zien, N.~Kraemer, S.~Sonnenburg, and G.~Raetsch.
\newblock {The Feature Importance Ranking Measure}.
\newblock In {\em ECML PKDD}, number~1, pages 1--15, 6 2009.

\end{thebibliography}
	\end{document}